\documentclass[10pt,letterpaper]{article}

\usepackage{cogsci}
\usepackage{pslatex}
\usepackage{apacite}
\usepackage{graphicx}

\usepackage[justification=centering]{caption}
\usepackage{array}
\usepackage{ragged2e}
\usepackage{epstopdf}

\usepackage{amsmath}%
\usepackage{amsfonts}%
\usepackage{amssymb}%

\title{A Computational Model of Two Cognitive Transitions Underlying Cultural Evolution}
 
\author{{\large \bf Liane Gabora (liane.gabora@ubc.ca)}\\
  University of British Columbia\\
  Department of Psychology, Okanagan campus, Arts Building, 3333 University Way\\
  Kelowna BC, V1V 1V7, CANADA\\
 \AND {\large \bf Wei Wen Chia (cww9989@gmail.com) and Hadi Firouzi (hadi.firouzi@ubc.ca)}\\
  University of British Columbia \\
  Department of Engineering, 5000-2332 Main Mall\\
  Vancouver BC,V6T 1Z4, CANADA\\
}

\begin{document}

\maketitle

 \noindent {Reference: Gabora, L., Chia, W. W., \& Firouzi, H. (2013). A computational model of two cognitive transitions underlying cultural evolution. {\it Proceedings of the Annual Meeting of the Cognitive Science Society}. July 31-3, Berlin. Austin TX: Cognitive Science Society.}\\
 
\begin{abstract}
\vskip -0.1in
We tested the computational feasibility of the proposal that open-ended cultural evolution was made possible by two cognitive transitions: (1) onset of the capacity to chain thoughts together, followed by (2) onset of contextual focus (CF): the capacity to shift between a divergent mode of thought conducive to `breaking out of a rut' and a convergent mode of thought conducive to minor modifications. These transitions were simulated in EVOC, an agent-based model of cultural evolution, in which the fitness of agents' actions increases as agents invent ideas for new actions, and imitate the fittest of their neighbors' actions. Both mean fitness and diversity of actions across the society increased with chaining, and even more so with CF, as hypothesized. CF was only effective when the fitness function changed, which supports its hypothesized role in generating and refining ideas.
% Because CF facilitated the generation of highly novel actions, it proved particularly effective when a new fitness function was introduced.  

\vskip -1in
\end{abstract}

\section{Introduction}
\vskip -0.05in
% It has been proposed that what is at the core of our uniquely human cognitive abilities is the capacity to place things in context, or see things from different perspectives (Gabora, 2003, 2008; Gabora \& Russon, 2011; Gabora \& Sternberg, 2010). This enables us to be not just creative, but to put our own spin on the inventions of others, modifying them to suit our own needs and tastes, in turn leading to new innovations that build cumulatively on previous ones. It enables us to modify ideas, attitudes and elements of knowledge by thinking about them in different contexts, and thereby weave them into a more or less integrated structure that characterizes who we are in relation to the world. 

Humans are unique with respect to the ability to generate accumulative, adaptive cultural evolution, a phenomenon referred to as the ratchet effect (Tomasello, Kruger, \& Ratner, 1993). Gaining insight into the origins of the capacity for complex culture is difficult, since all that is left of our prehistoric ancestors are bones and artifacts such as stone tools that resist the passage of time. Although methods for analyzing these remains are becoming increasingly sophisticated, they cannot always distinguish amongst competing theories. Thus, formal models provide valuable reconstructive tools for testing the feasibility of theories concerning the origins of the cognitive mechanisms that have transformed our planet. 

Several cognitive mechanisms have been implicated in the ability to evolve culture. One is the capacity to {\it chain} thoughts together to generate a sequence of actions or stream of thought (Donald, 1991). Another is {\it contextual focus} (hereafter referred to as CF): the capacity to shift between analytic and associative modes of thought (Gabora, 2003). Mathematical models of both have been developed (Gabora \& Aerts, 2009; Gabora \& Kitto, 2012; Veloz \emph{et al.}, 2011). Incorporating chaining into a computational model of cultural evolution increased the fitness and diversity of cultural outputs, as well as the effectiveness of learning (Gabora \& Saberi, 2011). Incorporating CF into a portrait painting computer program generated artworks that humans preferred over those generated without CF (DiPaola \& Gabora, 2009). However, the portrait painting program did not allow investigation of the effect of CF on the evolution of ideas through cultural interaction. The goal of the work presented here was to understand the relationship between chaining and CF. Specifically, we investigate the feasibility of the hypothesis that RR is broadly useful for improving cultural outputs, while CF is specifically useful for overcoming a new or sudden challenge. 

\subsection{Early Signs of Human Creativity}
\vskip -0.05in
The minds of our earliest ancestors, \emph{Homo habilis}, are referred to as {\it episodic} because there is no evidence that their experience deviated from the present moment of concrete sensory perceptions (Donald, 1991). They encoded perceptions of events in memory, but had little voluntary access to them without cues. 
% They were unlikely to think of a particular object, for example, unless something in their environment triggered it. 
They were therefore unable to voluntarily shape, modify, or practice skills and actions, and could not invent or refine complex actions, gestures, or vocalizations.

\emph{Homo habilis} was eventually replaced by \emph{Homo erectus}, which lived between approximately 1.8 and 0.3 million years ago. This period is considered the beginning of human cultural evolution. \emph{Homo erectus} exhibited signs of enhanced intelligence, creativity, and adaptability. They made sophisticated task-specific stone hand axes, had complex stable seasonal home bases, and there is evidence of long-distance hunting strategies involving large game, and migration out of Africa (Leakey 1984). It is widely believed that these early signs of creative culture reflect an underlying transition in cognitive or social abilities. 
The cranial capacity of the Homo erectus brain was approximately 1,000 cc, which is about 25\% larger than that of \emph{Homo habilis}, and at least twice as large as that of living great apes, and 75\% that of modern humans (Aiello, 1996
% ; Ruff et al, 1997
). 

Some have suggested that these abilities are due to the onset of a theory of mind (Mithen, 1998) or the capacity to imitate (Dugatkin, 2001). However, there is evidence that nonhuman primates also possess theory of mind (Heyes, 1998) 
%; Premack, 1988
and the capacity to imitate (Dugatkin, 2001), yet their cultural complexity do not compare with humans'. Evolutionary psychologists have suggested that our unique abilities were due to the onset of massive modularity 
% Buss, 1999, 2004; 
(Barkow, Cosmides, \& Tooby, 1992). However, although the mind exhibits an intermediate degree of functional and anatomical modularity, neuroscience has not revealed vast numbers of hardwired, encapsulated, task-specific modules; indeed, the brain is more subject to environmental influence than was previously believed (Buller, 2005; Byrne, 2000; Wexler, 2006).

Donald (1991) proposed that with the enlarged cranial capacity of \emph{Homo erectus}, the human mind underwent a transition characterized by a shift from an \emph{episodic} to a \emph{mimetic} mode of cognitive functioning, made possible by onset of the capacity to voluntarily retrieve memories independent of environmental cues and chain them into sequences. Donald refers to the cognitive architecture underlying this capacity as a \emph{self-triggered recall and rehearsal loop}. It enabled information to be processed recursively, and from different perspectives. Voluntary access to memories made it possible to act out\footnote{The term mimetic is derived from ``mime,'' which means ``to act out.''} events that occurred in the past or that might occur in the future. Thus not only could the mimetic mind temporarily escape the here and now, but by miming or gesture it could communicate similar escapes to other minds. The capacity to mime thus brought forth what is referred to as a \emph{mimetic} form of cognition, and allowed for the onset of culture. The self-triggered recall and rehearsal loop also enabled our ancestors to engage in a stream of thought, in which one thought or idea evokes another, and so forth recursively. In this way, attention can be directed away from the external world toward one's internal model of it. Finally, self-triggered recall allowed for voluntary rehearsal and refinement of actions, enabling systematic evaluation and improvement of skills and motor acts.

\subsection{An Explosion of Creative Cultural Change}
\vskip -0.05in
The European archaeological record indicates that an unparalleled cultural transition occurred between 60,000 and 30,000 years ago, at the onset of the Upper Paleolithic. Considering it "evidence of the modern human mind at work," Leakey (1984:93-94) describes this period as "unlike previous eras, when stasis dominated, ... [with] change being measured in millennia rather than hundreds of millennia." Similarly, Mithen (1998) refers to the Upper Paleaolithic as the `big bang' of human culture, exhibiting more innovation than in the previous six million years of human evolution. It marks the beginnings of traits considered diagnostic of behavioral modernity, including a more organized, strategic, season-specific style of hunting involving specific animals at specific sites, elaborate burial sites indicative of ritual and religion, evidence of dance, magic, and totemism, colonization of Australia, and replacement of Levallois tool technology by blade cores in the Near East. In Europe, complex hearths and many forms of art appeared, including cave paintings of animals, decorated tools and pottery, bone and antler tools with engraved designs, ivory statues of animals and sea shells, and personal decoration such as beads, pendants, and perforated animal teeth, many of which may have indicated social status.
%  (White, 1989)

Whether this period was a genuine revolution culminating in behavioral modernity is hotly debated because claims to this effect are based on the European Palaeolithic record, and largely exclude the African record (McBrearty \& Brooks, 2000).
% Henshilwood \& Marean, 2003). Many artifacts associated with a rapid transition to behavioral modernity 40-50,000 years ago in Europe are found in the African Middle Stone Age 
%tens of thousands of years earlier. 
However the 
% traditional and 
dominant view is that modern behavior appeared in Africa between 40,000 and 50,000 years ago, and spread, resulting in displacement of the Neanderthals in Europe
% Ambrose, 1998; Gamble, 1994; Klein, 2003; Stringer \& Gamble, 1993
(Klein, 1999). From this point on there was only one hominid species: modern \emph{Homo sapien}, and despite a lack of overall increase in cranial capacity, their prefrontal cortex, and more particularly the orbitofrontal region, increased significantly in size
% Deacon, 1997; 
(Dunbar, 1993).
%; Jerison, 1973; Krasnegor, Lyon, and Goldman-Rakic, 1997; Rumbaugh, 1997) 
in what was most likely a time of major neural reorganization (Klein, 1999). 
%; Henshilwood, d'Errico, Vanhaeren, van Niekerk, \& Jacobs, 2000; 
%Pinker, 2002
Given that the Middle/Upper Palaeolithic was a period of unprecedented creativity, what kind of cognitive processes were involved?

It is widely believed that a divergent or associative mode of thought predominates during idea generation, while a convergent or analytic mode predominates during the refinement, implementation, and testing of an idea (Finke, Ward, \& Smith, 1992). 
% ; Howard-Jones \& Murray, 2003; Martindale, 1995; Smith, Ward, \& Finke, 1995; Ward, Smith, \& Finke, 1999
It has been proposed that the Paleolithic transition reflects fine-tuning of the biochemical mechanisms underlying the capacity to subconsciously shift between these modes, depending on the situation, by varying the specificity of the activated cognitive receptive field (Gabora, 2003; Gabora \& Kaufman, 2010). This is referred to as contextual focus\footnote{In neural net terms, CF amounts to the capacity to spontaneously and subconsciously vary the shape of the activation function, flat for divergent thought and spiky for analytical} (CF) because it requires the ability to focus or defocus attention in response to the context or situation one is in. Defocused attention, by diffusely activating a broad region of memory, is conducive to divergent thought; it enables obscure (but potentially relevant) aspects of the situation to come into play. Focused attention is conducive to convergent thought; memory activation is constrained enough to hone in and perform logical mental operations on the most clearly relevant aspects. 
% The theory is consistent with the widespread belief that the generation of cultural novelty involves both freedom and constraint.

\section{The Computational Model}
We reviewed the evidence for two hypotheses: (1) the earliest signs of culture were due to the onset of the capacity to chain representations together, and (2) the cultural explosion of the Middle-Upper Paleolithic was due to the onset of CF. We investigated these hypotheses using an agent-based model of cultural evolution referred to as ``EVOlution of Culture'', abbreviated EVOC. EVOC uses neural network based agents that (1) invent new ideas, (2) imitate actions implemented by neighbors, (3) evaluate ideas, and (4) implement successful ideas as actions. EVOC is an elaboration of Meme and Variations, or MAV (Gabora, 1995), the earliest computer program to our knowledge to model not just cultural transmission but cumulative, adaptive, cultural evolution.\footnote{The approach can thus be contrasted with computer models of how individual learning affects biological evolution (e.g., Higgs, 2000; Hinton \& Nowlan, 1987; Hutchins \& Hazelhurst, 1991).} It was inspired by the genetic algorithm, a search technique that finds solutions to complex problems by generating a `population' of candidate solutions through processes akin to mutation, selecting the best, and repeating until a satisfactory solution is found (Holland, 1975). The goal behind MAV, and also behind EVOC, was to distil the underlying logic of not biological evolution but cultural evolution. Agents do not evolve in a biological sense--they neither die nor have offspring--but do in a cultural sense, by adaptively modifying each others' ideas for actions. 
%In cultural evolution, the generation of novelty takes place through invention instead of through mutation and recombination as in biological evolution, and the differential replication of novelty takes place through imitation, instead of through reproduction with inheritance as in biological evolution. EVOC has been used to address such questions as how does the presence of leaders or barriers to the diffusion of ideas affect cultural evolution. 
We summarize the architecture of EVOC in sufficient detail to explain our results; for details we refer the reader to previous publications ({\it e.g.,} Gabora, 1995; Gabora \& Saberi, 2011; Leijnen \& Gabora, 2009).

\subsection{Agents}
\vskip -0.05in
Agents consist of (1) a neural network, which encodes ideas for actions and detects trends in what constitutes a fit action, (2) a `perceptual system', which carries out the evaluation and imitation of neighbours' actions, and (3) a body, consisting of six body parts which implement actions. The neural network is composed of six input nodes and six corresponding output nodes that represent concepts of body parts (LEFT ARM, RIGHT ARM, LEFT LEG, RIGHT LEG, HEAD, and HIPS), as well as hidden nodes that represent more abstract concepts (LEFT, RIGHT, ARM, LEG, SYMMETRY, OPPOSITE, and MOVEMENT). Input nodes and output nodes are connected to hidden nodes of which they are instances (\emph{e.g.,} LEFT ARM is connected to LEFT.) Activation of any input node activates the MOVEMENT node. Same-direction activation of symmetrical input nodes (\emph{e.g.,} upward motion--of both arms) activates the SYMMETRY node. 
% In the experiments reported here the OPPOSITE hidden node was not used.

\subsection{Invention}
\vskip -0.05in
An idea for a new action is a pattern consisting of six elements that dictate the placement of the six body parts. Agents generate new actions by modifying their initial action or an action that has been invented previously or acquired through imitation. During invention, the pattern of activation on the output nodes is fed back to the input nodes, and invention is biased according to the activations of the SYMMETRY and MOVEMENT hidden nodes. (Were this not the case there would be no benefit to using a neural network.) To invent a new idea, for each node of the idea currently represented on the input layer of the neural network, the agent makes a probabilistic decision as to whether the position of that body part will change, and if it does, the direction of change is stochastically biased according to the learning rate. If the new idea has a higher fitness than the currently implemented idea, the agent learns and implements the action specified by that idea. 

\subsection{Imitation}
\vskip -0.05in
The process of finding a neighbour to imitate works through a form of lazy (non-greedy) search. The imitating agent randomly scans its neighbours, and adopts the first action that is fitter than the action it is currently implementing. If it does not find a neighbour that is executing a fitter action than its own current action, it continues to execute the current action. 

%added 2nd sentence -LC
\subsection{Evaluation: The Fitness Function}
\vskip -0.01in
Fitness was evaluated using an adaptation of the Royal Roads fitness function (Forrest \& Mitchell, 1993). Midway through a run the fitness function was changed to test the effectiveness of chaining and CF for adapting to a sudden change in the task constraints or the environment. Definitions of terms used to accomplish this are provided in Table One. 

\begin{table}
\caption{Definition table.}
\label{fig:Definition table.}
%\vskip 0.12in
\begin{tabular}{|p{0.07\textwidth}|p{0.2\textwidth}|p{0.13\textwidth}|}
	\hline
	\textbf{Term} & \textbf{Definition} & \textbf{Example} \\	\hline
	Body Part	& Component of agent other than neural network.	& Left Arm (LA) \\ \hline
	Sub-action & Set of six components that indicates position of 6 body parts. Each can be in a neutral (0), up (1), or down (-1) position.  &  {HD:0, LA:1, RA:-1, LL:1, RL:0, HP:-1; This sub-action is abbreviated 01-110-1} \\ \hline
Action & One or more sequential sub-actions. & {{01001-1}, {-10-1-111}}  \\ \hline
Template	& Abstract or prototypical format for a sub-action. Position of a body part can be unspecified (*). & {HD:0, LA:*, RA:1, LL:*, RL:1, HP:-1} \\ \hline
% Diversity & Number of different actions & \\ \hline
\end{tabular}
% \vskip -.1in
\end{table}

%edited first sentence -LC
The first fitness function is determined by 45 templates, six of which are shown in Table Two. The second (not shown) is constructed analogously, with different sub-actions. The templates can be thought of as defining the cultural significance of types of sub-actions (such as dance steps). Each template $T^i$ consists of six components, one for each body part (\emph{i.e.,} $T^i={t_j^i };j=1..6$). Each body part can be in a neutral position (0) , up (1), down (-1), or an unspecified position (*). For example, in template $T^i={*,1,-1,*,*,0}$, the left arm is up (LA:1), the right arm is down (RA:-1), the hips are in the neutral position (HP:0), and the positions of other body parts is unspecified (HD:*, LL:*, and RL:*). The templates provide constraints, as well as flexibility with respect to what constitutes a fit action. For example, in an optimally fit action, the head must be in the neutral position (in $T^1$ the first component is 0) but the positions of other body parts can vary). 
% Also, the arms and legs must be in opposite positions (e.g., left arm/leg up and right arm/leg down, or \emph{vice versa}). Moreover, the left arm and left leg (right arm and right leg) must be in the same position. For example in $T^{16}$, both the left arm and left leg are up '1' and others are unimportant '*'.  As a result, four optimal sub-actions can be found when the head is neutral, the left arm and left leg are either up or down, the right arm and leg are down (up), and the head is up (down). 
The optimal sub-actions are $\{0,1,-1,1,-1,1\}$, $\{0,1,-1,1,-1,-1\}$, $\{0,-1,1,-1,1,1\}$, and $\{0,-1,1,-1,1,-1\}$.

%first fitness function templates -LC
\begin{table}% [!ht]
 \begin{center} 
 \caption{Partial set of the templates used in the first fitness function. (The rest are omitted due to lack of space.)} 
\label{A subset of the set of templates used in the first fitness function} 
%\vskip 0.12in
\begin{tabular}{|c|c|c|}
\hline
$T^1=\{0,*,*,*,*,*\}$ & $T^{24}=\{1,*,*,1,1,*\}$  \\ \hline
$T^2=\{*,0,*,*,*,*\}$ & $T^{25}=\{1,*,1,*,1,*\}$   \\ \hline
$T^3=\{*,*,0,*,*,*\}$ & $T^{26}=\{1,*,1,1,*,*\}$  \\ \hline
% $T^4=\{*,*,*,0,*,*\}$ & $T^{27}=\{*,-1,1,-1,*,*\}$  \\ \hline
% $T^5=\{*,*,*,*,0,*\}$ & $T^{28}=\{*,-1,1,*,1,*\}$  \\ \hline
% $T^6=\{*,*,*,*,*,0\}$ & $T^{29}=\{1,-1,1,*,*,*\}$ \\ \hline
% $T^7=\{*,-1,-1,*,*,*\}$ & $T^{30}=\{*,-1,*,-1,1,*\}$ \\ \hline
% $T^8=\{*,*,-1,-1,*,*\}$ & $T^{31}=\{1,-1,*,-1,*,*\}$ \\ \hline
% $T^9=\{-1,*,*,-1,*,*\}$ & $T^{32}=\{1,-1,*,*,1,*\}$  \\ \hline
% $T^{10}=\{*,-1,*,-1,*,*\}$ & $T^{33}=\{*,*,1,-1,1,*\}$ \\ \hline
% $T^{11}=\{-1,-1,*,*,*,*\}$ & $T^{34}=\{1,*,*,-1,1,*\}$  \\ \hline
% $T^{12}=\{-1,*,-1,*,*,*\}$ & $T^{35}=\{1,*,1,-1,*,*\}$   \\ \hline
% $T^{13}=\{-1,*,*,*,-1,*\}$ & $T^{36}=\{*,1,-1,-1,*,*\}$  \\ \hline
% $T^{14}=\{*,-1,*,*,-1,*\}$ & $T^{37}=\{*,1,-1,*,-1,*\}$  \\ \hline
% $T^{15}=\{*,*,-1,*,-1,*\}$ & $T^{38}=\{1,1,-1,*,*,*\}$  \\ \hline
% $T^{16}=\{*,*,*,-1,-1,*\}$ & $T^{39}=\{*,1,*,-1,-1,*\}$ \\ \hline
% $T^{17}=\{*,1,1,1,*,*\}$ & $T^{40}=\{1,1,*,-1,*,*\}$ \\ \hline
% $T^{18}=\{*,1,1,*,1,*\}$ & $T^{41}=\{1,1,*,*,-1,*\}$ \\ \hline
% $T^{19}=\{1,1,1,*,*,*\}$ & $T^{42}=\{*,*,-1,-1,-1,*\}$  \\ \hline
% $T^{20}=\{*,1,*,1,1,*\}$ & $T^{43}=\{1,*,*,-1,-1,*\}$ \\ \hline
% $T^{21}=\{1,1,*,1,*,*\}$ & $T^{44}=\{1,*,-1,*,-1,*\}$  \\ \hline
% $T^{22}=\{1,1,*,*,1,*\}$ & $T^{45}=\{1,*,-1,-1,*,*\}$  \\ \hline
% $T^{23}=\{*,*,1,1,1,*\}$ &\\ \hline
\end{tabular} 
 \end{center} 
 \vskip -.15in
\end{table}

Assume that D is a sub-action (\emph{i.e.,} $D={d_j};j=1..6$) and $T^i$ is the $i^{th}$ template (\emph{i.e.,} $T^i={t_j^i};j=1..6$). Thus, $d_j$ represents the position of the $j^{th}$ body part and the value of $d_j$ can be either 0 (neutral), 1 (up), or -1 (down). Likewise, the value of $t_j^i$ can be 0, 1, -1, or * (unspecified). Accordingly, the fitness of sub-action D is obtained as per Eq. \ref{eq:F D}.

\begin{equation}
	F(D)=\sum_{i=1}^{19}{\Phi(T^i,D) \times \Omega(T^i)}
\label{eq:F D}
\end{equation}

As shown in Eq.~\ref{eq:F D}, fitness is a function of template weight ($\Phi(T^i,D)$) and template order ($\Omega(T^i)$).

\subsubsection{Template Weight}
\vskip -0.05in
$\Phi(T^i,D)$ is a function that determines the weight of sub-action $D$ by comparing it with template $T^i$. This weight is set to one if each component of the sub-action (\emph{i.e.,} $d_j;j=1..6$) either matches the corresponding component of the template (\emph{i.e.,} $t_j^i;j=1..6$) or if  the corresponding components of the template is unspecified (\emph{i.e.,} $t_j^i= *$):
\vskip -.1in
\begin{eqnarray}
	\Phi(T^i,D) = \bigg\{ 
	\begin{matrix}
		1 & if \ \forall t_j^i \in T^i:t_j^i= d_j \ or \ * \\
		0 & otherwise	
	\end{matrix}	
\label{eq:template weight}
\end{eqnarray}

\subsubsection{Template Order}
\vskip -0.05in
$\Omega(T^i)$ computes the order of the template $T^i$ by counting the number of components that have a specified value (\emph{i.e.,} $t_j^i \neq *)$.

\begin{equation}
\Omega(T^i)= \sum_{j=1, t_j^i \neq *}^6{t_j^i}
\label{eq:template order}
\end{equation}

%replaced original text below -LC
The fitness functions are difficult to solve because they are rugged; is to have multiple milestones, or fitness peaks, that agents must achieve before reaching the plateau. For example, consider the fitness function given in Table 2. The action {0,0,0,0,0,0} has a fitness of 6. An agent may move on from this action to find an actions that fits the third order templates with a fitness of 31, {\it e.g.,} $F(D):\{1,1,1,1,1,0\}=3+3+3+3+3+3+3+3+3+3+1=31$. 

\subsection{Learning}
\vskip -0.05in
Invention makes use of the ability to learn, and respond adaptively to trends. 
% Since no action acquired through imitation or invention is implemented unless it is fitter than the current action, new actions provide information about what constitutes a fit idea. 
Knowledge acquired through the evaluation of actions is translated into educated guesses about how to invent fit actions. For example, an agent may learn that symmetrical movement tends to be either beneficial or detrimental, and bias the generation of new actions accordingly.

\subsection{A Typical Run}
\vskip -0.05in
Fitness and diversity of actions are initially low because all agents are initially immobile, implementing the same action, with all body parts in the neutral position. Soon some agent invents an action that has a higher fitness than immobility, and this action gets imitated, so fitness increases. Fitness increases further as other ideas get invented, assessed, implemented as actions, and spread through imitation. The diversity of actions increases due to the proliferation of new ideas, and then decreases as agents hone in on the fittest actions. Thus, over successive rounds of invention and imitation, the agents' actions improve. EVOC thereby models how ``descent with modification'' occurs in a purely cultural context. 

\section{Method}
%\vskip -0.005in
\subsection{Modeling Chaining}
\vskip -0.01in
The chaining algorithm is illustrated schematically in Figure 1b. Chaining gives agents the opportunity to execute multi-step actions. The agent can keep adding a new sub-action to its current action so long as the most recently-added sub-action is both novel and successful. A sub-action D is considered novel if at least one of its components is different from that of the previous sub-action. It is considered successful if there exists a template $T^i$ such that $\Phi(T^i,D)$ is one. 
% This seems to be a common feature of many useful actions such as the repetitive motions involved in tool-making, sawing, carving, weaving, and so forth. 
\vskip -.1in
\begin{equation}
successful(D) = \bigg\{ \begin{matrix}
	true & if \ \exists \ T^i : \Phi(T^i,D) = 1 \\
	false & otherwise
\end{matrix}
\label{eq:chaining}
\end{equation}

The fitness of an action consisting of more than one sub-action is obtained by adding the number of sub-actions to the fitness of the last sub-action in the sequence. For example, if the last sub-action of an action is $D=[0,1,-1,1,-1,1]$ and the number of sub-actions is seven, the fitness of the action is $F(D)+7=14+7=21$. Thus where $c$ is `with chaining', $w$ is `without chaining', n is the number of chained sub-actions, the fitness of a chained action, $F_c$, is calculated as follows:

\begin{equation}
	F_c=F_w+n
\label{eq:chaining fitness}
\end{equation}

An agent can execute an arbitrarily long action so long as it continues to invent successful new sub-actions. In general, the more sub-actions the fitter the action. This is admittedly a simple algorithm of simulating the capacity for chaining, but we were not interested in the impact of this action {\it per se}. The goal here was simply to test hypotheses about how chaining at the individual level affects dynamics at the societal level, by providing agents with a means of implementing multistep actions such that the optimal way of going about one step depends on how one went about the previous step. 

 \subsection{Modeling Contextual Focus}
%\vskip -0.01in
% This is the old version of this section: 
% The CF algorithm is illustrated schematically in Figure 1c. In the convergent mode, the current action is only slightly modified to create a new action. In the divergent mode, the current action is substantial modified to create a new action. An agent switches between these modes by modifying its {\it rate of conceptual change}. If the fitness of an action suddenly decreases relative to previous actions, the rate of conceptual change increases, the agent shifts to a more divergent processing mode of thought, conducive to large leaps through the space of possibilities. When the fitness of an action is high relative to previous actions, the rate of conceptual change decreases, and the agent shifts to a convergent mode conducive to minor adjustments. When CF is off, the rate of conceptual change stays constant throughout the run. 

The CF algorithm is illustrated schematically in Figure 1c. In the convergent mode, the current action is only slightly modified to create a new action. In the divergent mode, the current action is substantial modified to create a new action. An agent switches between these modes by modifying its {\it rate of creative change} (RCC). If the fitness of its current action is low relative to previous actions, RCC increases, causing the agent to shift to a more divergent processing mode conducive to large leaps through the space of possibilities. If action fitness is high relative to that of previous actions, RCC decreases, and the agent shifts to a more convergent mode conducive to minor adjustments. With CF turned off, RCC stays constant throughout the run at 1/6 ({\it i.e.,} a new action involves change to one of the six body parts). The equation to modify RCC is shown in Eq. 6 where $a$ is a negative value. Since at the start of a run previous fitness is undefined, RCC in this case is a function of the current fitness as per Eq. 7, where  0 $< b < 1$. 

%CF delta rcc equation (i am not sure how the equation below will turn out) -LC
\begin{equation}
	\Delta RCC=a(F_{new}-F_{old})
\label{eq:delta rcc}
\end{equation}
%equation to initialise RCC -LC
\begin{equation}
	RCC_{initial}=b^{F_{current}}
\label{eq:initial rcc}
\end{equation}

%paragraph added below to discuss CF equations -LC
In the results shown here $a$ and $b$ were initialized to -0.005 and 0.8 respectively.

\begin{figure}%
\centering
\includegraphics[width=0.95\columnwidth]{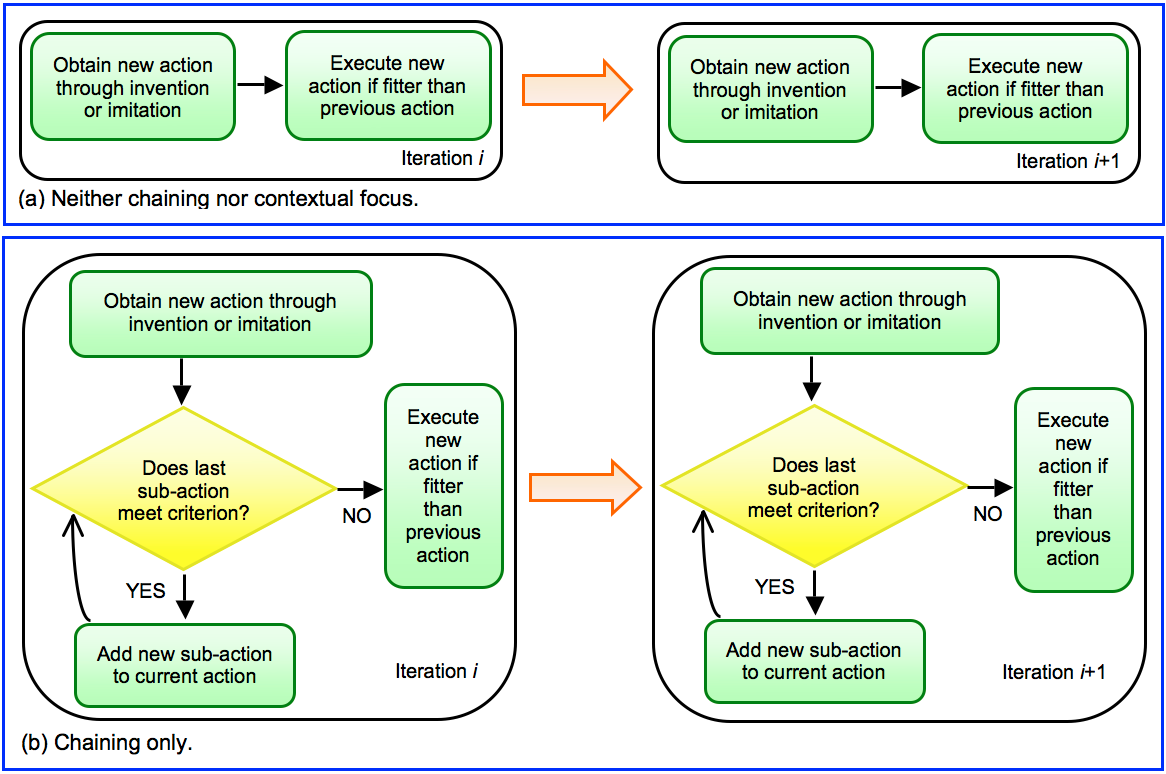} \\
\vskip 0.01in
\includegraphics[width=0.95\columnwidth]{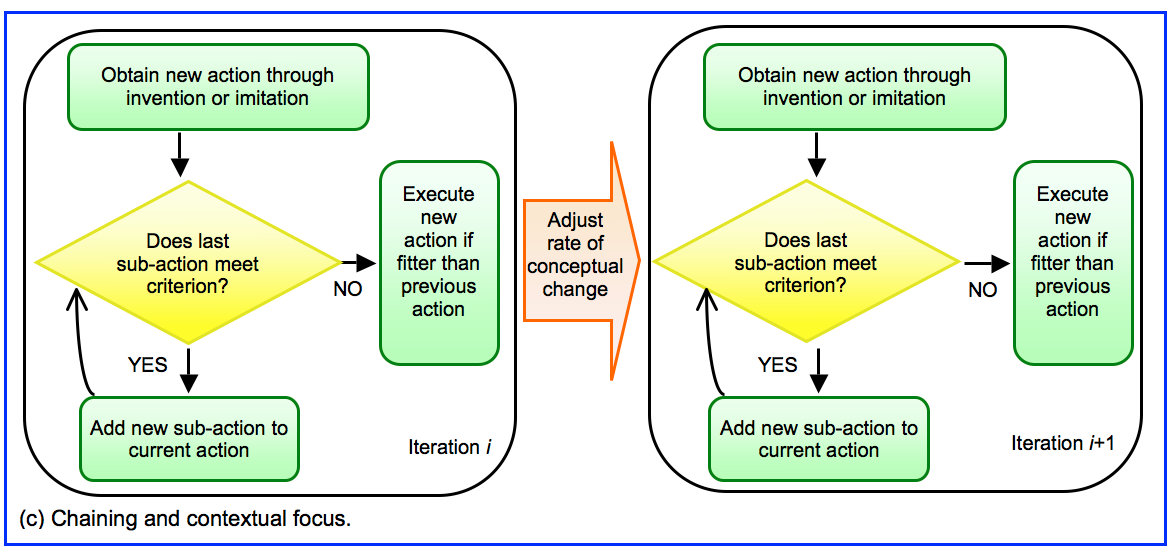}
 \begin{center} 
\caption{Schematic illustration of (a) neither chaining nor CF, (b) chaining only, and (c) both. Chaining operates within a generation whereas CF operates between generations. 
% See text for details.
}%
 \end{center} 
\label{fig:Schematic illustration of the CF algorithm.}%
\vskip -0.3in
\end{figure}

\section{Results}
\vskip -0.03in
The results of introducing chaining and CF on the mean fitness and diversity (total number of different actions) of actions across all agents in the society are shown in Figures 2 and 3 respectively. All graphs show means of 500 runs. Chaining and CF both significantly increased mean fitness of actions. Without chaining, mean fitness quickly reached a plateau; with chaining it could increase indefinitely. While chaining increased mean fitness throughout the run, CF only increased mean fitness following initial exposure to a new fitness function, i.e., at the beginning of the run, and when the second fitness function was introduced at iteration 50.

\begin{figure}[ht]
\centering
\includegraphics[width=0.95\columnwidth]{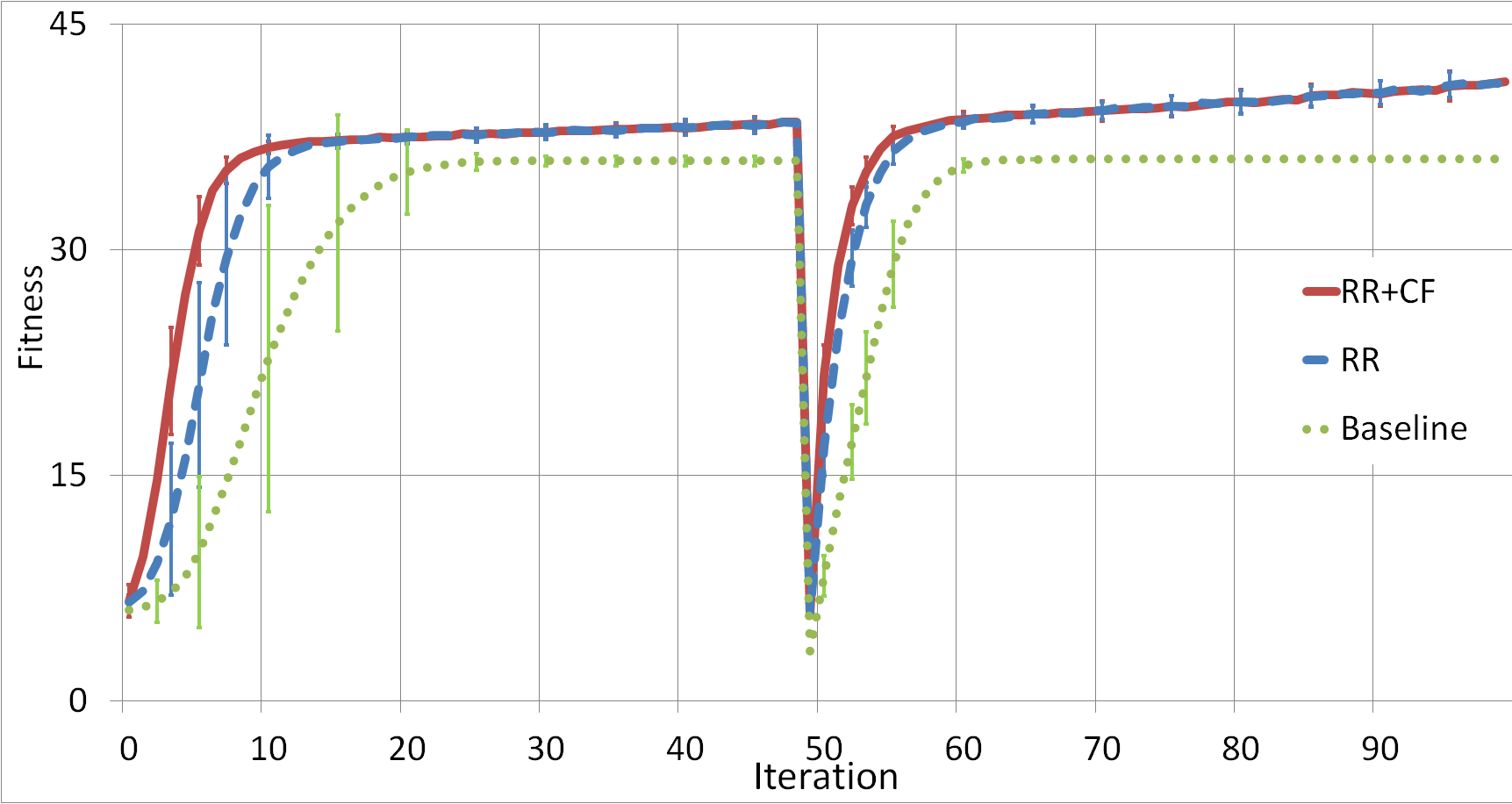}
\caption{Mean fitness of cultural outputs across the society with both chaining and CF (red line), chaining only (dashed blue line), and neither chaining nor CF (dotted green line).}
\label{fig:fig-fitness}
\vskip -0.1in
\end{figure}

Chaining also significantly increased the diversity of actions. Although inspection reveals that there is always convergence on optimal actions, without chained actions, this set is a static (thus mean fitness plateaus) whereas with chained actions the set of optimal actions changes, as increasingly fit actions continue to be found. When agents were first exposed to a fitness function, CF increased both the rate at which new possibilities were generated, and the rate of convergence on the fittest of these, although this effect is more pronounced for the first fitness function than the second. As with fitness, CF exerted no noticeable effect on diversity once the agents had fit actions. 

\begin{figure}[ht]
\centering
\includegraphics[width=0.95\columnwidth]{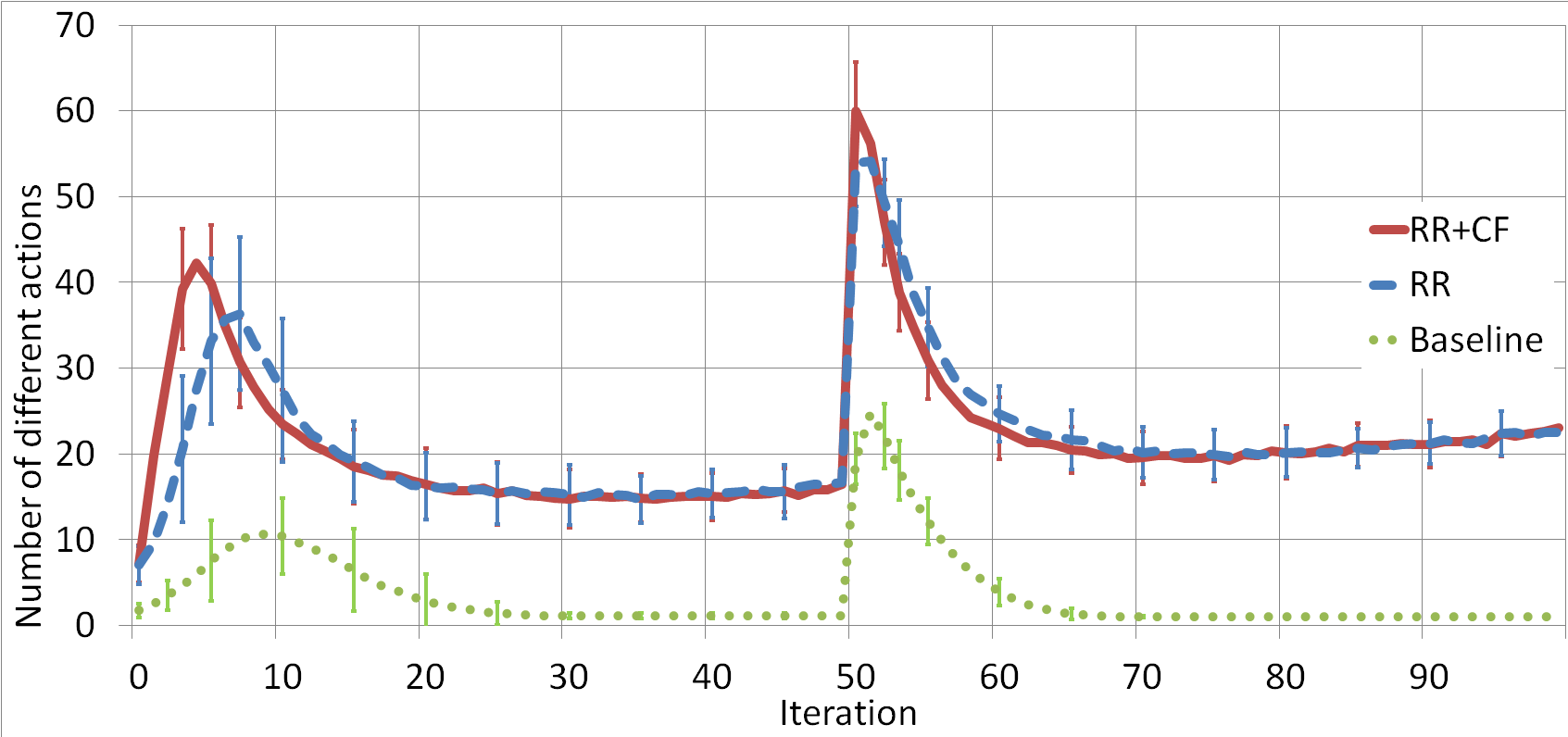}
\caption{Diversity of  cultural outputs across the society with both chaining and CF (red line), chaining only (dashed blue line), and neither chaining nor CF (dotted green line).}
\label{fig:fig-diversity}
\vskip -0.1in
\end{figure}
% \paragraph{Effect of CF}
% Fig~\ref{fig:mf_new_smp-ch-cf-chcf} shows the diversity (number of different actions) over time. Without chaining, the number of different actions is first increase to its maximum value and then converged to the number of optimal sub-actions (\emph{i.e.,} four). The number of different actions is always above four when chaining is used. CF also increases the diversity of actions.

\section{Discussion}
\vskip -0.01in
This paper provides valuable insights into the mechanisms underlying the uniquely human capacity for collectively generated, open-ended, adaptive cultural evolution. Our results suggest that once humans became able to sequence thoughts together to generate increasingly complex and refined cultural outputs, they would have found themselves at a significant adaptive advantage. Similarly, our results suggest that once humans became able to employ an exploratory, divergent processing mode when stuck, followed by a shift to a more constrained convergent processing mode to fine-tune their cultural outputs, they would have been capable of generating significantly more valuable cultural outputs. We suggest that a mechanism akin to CF is what makes possible the cumulative creativity exhibited by successful computational models of language evolution (\emph{e.g.,} Kirby, 2001). A potential downfall of processing in a divergent mode is that since effort is devoted to the re-processing of previously learned material, less effort may be devoted to being on the lookout for danger and simply carrying out practical tasks. Since divergent thought carries a high cognitive load, it would not have been useful to think divergently until there was a means to shift back to a convergent mode. Although these results do not prove that onset of the capacity to chain thoughts together into sequences, and to shift between divergent modes of thought through CF, are responsible for our cultural complexity, it shows that they provide a computationally feasible explanation. We know of no other cognitive mechanisms implicated in the evolution of complex culture for which open-ended, adaptive cultural change has been demonstrated.  

Both chaining and CF were implemented in a simple manner. Future investigations will focus on developing more realistic implementations of chaining and CF. Chaining will use associative recall to reconsider an item from multiple potentially relevant `perspectives', and the divergent mode of CF will use a sophisticated mathematical model of concepts to facilitate the generation of new concept combinations.

\section{Acknowledgments}
This project was conducted at the Media and Graphics Interdisciplinary Centre (MAGIC), UBC. It was supported by the Natural Sciences and Engineering Research Council of Canada, and Fund for Scientific Research, Flanders, Belgium.

%\vspace{-0.2cm}
% \small
\section*{References}
\begin{description}
 \setlength{\itemsep}{-1mm}

% Aerts, D. (2009). Quantum structure in cognition. Journal of Mathematical Psychology, 53, 314 348.

% Aerts, D., & Gabora, L. (2005). A state-context-property model of concepts and their combinations I: The structure of the sets of contexts and properties. Kybernetes, 34(1&2), 167-191. 

% Aerts, D., & Gabora, L. (2005). A state-context-property model of concepts and their combinations II: A Hilbert space representation. Kybernetes, 34(1&2), 192-221.

\item Aiello, L. C. (1996). Hominine pre-adaptations for language and cognition. In P. Mellars \& K. Gibson (Eds.), {\it Modeling the early human mind}, Cambridge, UK: McDonald Institute Monographs, 89-99.

% Bak, P., Tang, C.  \& Weisenfeld, K. (1988). Self-organized criticality. Physical Review A, 38, 364.

\item Barkow, J. H., Cosmides, L.,  \& Tooby, J., Eds. (1992). {\it The adapted mind 
% : Evolutionary psychology and the generation of culture.
} New York: Oxford University Press.

% Bentley, P. D.  \& Corne D., Eds. (2002). Creative evolutionary systems, San Francisco: Morgan Kaufmann.

% Bentley, R. A., Ormerod, P.,  \& Batty, M. (2011). Evolving social influence in large populations. Behavioral Ecology and Sociobiology, 65, 537-546.

% Best, M. (1999). How culture can guide evolution: An inquiry into gene/meme enhancement and opposition. Adaptive Behavior, 7(3), 289-293.

% Best, M. (2006). Adaptive value within natural language discourse. Interaction Studies, 7, 1-15.

% Brown, J. (2009). Looking at Darwin: portraits and the making of an icon. Isis, 100(3), 542-70.

% Bruza, P.D., Kitto, K., Nelson, D., McEvoy, C. (2009). Is there something quantum-like about the human mental lexicon? Journal of Mathematical Psychology, 53, 362-377. 

\item Buller, D. J. (2005). {\it Adapting minds.} MIT Press.

% \item Buss, D. M. (1999/2004). {\it Evolutionary psychology: The new science of the mind.} Boston, MA: Pearson.

\item Byrne, R.,  \& Russon, A. (1998). Learning by imitation: A hierarchical approach. {\it Behavioral and Brain Sciences, 2,} 667-721.

\item Cavalli-Sforza, L. L.,  \& Feldman, M. W. (1981). {\it Cultural transmission and evolution: A quantitative approach.} Princeton: Princeton University Press.

% Cloak, F. T. Jr. (1975). Is a cultural ethology possible? Human Ecology, 3, 161-182.

% DiPaola, S. & Gabora, L. (2007). Incorporating characteristics of human creativity into an evolutionary art algorithm. In (D. Thierens, Ed.), Proceedings of the Genetic and Evolutionary Computing Conference (pp. 2442-2449). University College London.

% DiPaola S. (2009). Exploring a Parameterized portrait painting space, International Journal of Art and Technology, 2(1-2), 82-93.

\item DiPaola, S.  \& Gabora, L. (2009). Incorporating characteristics of human creativity into an evolutionary art algorithm. {\it Genetic Programming  and Evolvable Machines, 10,} 97-110. 

%% \item Donald, M. (1998). Hominid enculturation and cognitive evolution. In C. Renfrew \& C. Scarre, Eds. {\it Cognition and material culture: The archaeology of symbolic storage} (pp. 7-17). Cambridge UK: McDonald Institute Monographs.

\item Donald, M. (1991). {\it Origins of the modern mind: Three stages in the evolution of culture and cognition.} Cambridge, MA: Harvard University Press.

\item Dugatkin, L. (2001). {\it Imitation factor: Imitation in animals and the origin of human culture.} New York: Free Press.

\item Dunbar, R. (1993). Coevolution of neocortical size, group size, and language in humans. {\it Behavioral and Brain Sciences, 16}, 681-735.

%% \item Gabora, L. (1994). A computer model of the evolution of culture. I{\it Proceedings of the 4th International Conference on Artificial Life.} Camgridge MA: MIT Press.
% Boston, MA.

\item Finke, R. A., Ward, T. B., \& Smith, S. M. (1992). {\it Creative cognition: Theory, research, and applications.} Cambridge, MA: MIT Press.

\item Forrest, S. \& Mitchell, M. (1993). Relative building block fitness and the building block hypothesis. In (L. Whitley, Ed.) {\it Foundations of genetic algorithms.} San Mateo: Morgan Kaufman. 

\item Gabora, L. (1995). Meme and Variations: A computer model of cultural evolution. In L. Nadel \& D. Stein (Eds.) {\it 1993 Lectures in Complex Systems.} Boston: Addison-Wesley.

%% \item Gabora, L. (1996). A day in the life of a meme. {\it Philosophica, 57,} 901-938. 

% Gabora, L. (1998). Autocatalytic closure in a cognitive system: A tentative scenario for the origin of culture. Psycoloquy, 9(67). http://www.cogsci.ecs.soton.ac.uk/cgi/psyc/newpsy?9.67

%% \item Gabora, L. (1999). Conceptual closure: Weaving memories into an interconnected worldview. In (G. Van de Vijver & J. Chandler, Eds.) {\it Closure: Emergent organizations and their dynamics.} University of Gent, Belgium: Research Community on Evolution and Complexity and  Washington Evolutionary Systems Society. 

% Gabora, L. (1999b). To imitate is human: A review of 'The Meme Machine' by Susan Blackmore. Journal of Artificial Societies and Social Systems 2(2). Reprinted in Journal of Consciousness Studies, 6, 77-81.

\item Gabora, L. (2003). Contextual focus: A cognitive explanation for the cultural transition of the Middle/Upper Paleolithic. {\it Proceedings Annual Conference Cognitive Science Society} (pp. 432--437), Boston: Lawrence Erlbaum.

%%\item Gabora, L. (2004). Ideas are not replicators but minds are. {\it Biology \& Philosophy, 19}(1), 127-143. 

%% \item Gabora, L. (2008a). Mind. In (R. A. Bentley, H. D. G. Maschner, & C. Chippindale, Eds.) {\it Handbook of theories and methods in archaeology,} Altamira Press, Walnut Creek CA, (pp. 283-296). 

%\item Gabora, L. (2008). EVOC: A computer model of the evolution of culture. {\it Proceedings 30th Annual Conference Cognitive Science Society} (pp. 1466--1471). Hanover PA: Sheridan.

%% \item Gabora, L. (2008c). Modeling cultural dynamics. {\it Proceedings of the Association for the Advancement of Artificial Intelligence (AAAI) Fall Symposium 1: Adaptive Agents in a Cultural Context} (pp. 18-25), AAAI Press.

%% \item Gabora, L. (2008b). The cultural evolution of socially situated cognition. {\it Cognitive Systems Research, 9}(1-2), 104-113.

\item Gabora, L. \& Aerts, D. (2009). A mathematical model of the emergence of an integrated worldview. {\it Journal of Mathematical Psychology, 53,} 434--451.

% Gabora, L. \& Leijnen, S. (2009). How creative should creators be to optimize the evolution of ideas? A computational model. Electronic Proceedings in Theoretical Computer Science, 9, 108-119.

 \item Gabora, L., \& Kaufman, S. (2010). Evolutionary perspectives on creativity. In (J. Kaufman \& R. Sternberg, Eds.) {\it The Cambridge handbook of creativity.} Cambridge: Cambridge University Press.

 \item Gabora, L., \& Kitto, K. (2012). Concept combination and the origins of complex cognition. In (L. Swan, Ed.) {\it Origins of mind.} Berlin: Springer.

%% \item Gabora, L. \& Russon, A. (2011). The evolution of human intelligence. In (R. Sternberg  \& S. Kaufman, Eds.) {\it The Cambridge handbook of intelligence.} Cambridge UK: Cambridge University Press.

\item Gabora, L. \& Saberi, M. (2011). How did human creativity arise? An agent-based model of the origin of cumulative open-ended cultural evolution. {\it Proceedings Conference on Cognition \& Creativity} (pp. 299-306). New York: ACM.
% Association for Computing Machinery.
% Atlanta, GA. 

%% \item Gabora, L., \& Firouzi, H. (2012). Society functions best with an intermediate level of creativity. {\it Proceedings of the Annual Meeting of the Cognitive Science Society} (pp. 1578-1583). August 1-4, Sapporo Japan.

% Hampton, J. (1987). Inheritance of attributes in natural concept conjunctions. Memory & Cognition, 15, 55-71.

% \item Hartley, J. (2009). From cultural studies to cultural science. {\it Cultural Science, 2,} 1-16.

\item Higgs, P. (2000). The mimetic transition: a simulation study of the evolution of learning by imitation. {\it Proceedings: Royal Society B: Biological Sciences, 267,} 1355-1361.

\item Heyes, C. M. (1998). Theory of mind in nonhuman primates. {\it Behavioral and Brain Sciences, 211,} 104-134.

\item Hinton, G. E.  \& Nowlan, S. J. (1987). How learning can guide evolution. {\it Complex Systems, 1,} 495-502.

\item Holland, J. K. (1975). {\it Adaptation in natural and artificial systems.} Ann Arbor, MI: University of Michigan Press.

\item Hutchins, E.  \& Hazelhurst, B. (1991). Learning in the cultural process. In Langton, C., 
% Taylor, J., 
Farmer, D., \& Rasmussen, S. (Eds.) {\it Artificial Life II.} Redwood City: Addison-Wesley.

% Isham, C. (1995). Lectures on quantum theory. London, UK. Imperial College Press.

\item Kirby, S. (2001). Spontaneous evolution of linguistic structure: An iterated learning model of the emergence of regularity and irregularity. {\it IEEE Transactions on Evolutionary Computation, 5}(2), 102-110.

% Kitto, K. (2006), Modelling and generating complex emergent behaviour, PhD thesis, School of Chemistry, Physics and Earth Sciences, The Flinders University of South Australia.

% Kitto, K. (2008a). High End Complexity, International Journal of General Systems, 37(6), 689-714.

% Kitto, K. (2008b).  Why Quantum Theory? Proceedings of the Second Quantum Interaction Symposium, pp. 11-18, College Publications. 

% Kitto, K., Ramm, B., Sitbon, L., Bruza, P. (2011). Quantum theory beyond the physical: information incontext. Axiomathes, 21(2):331-345.

% Kitto, K., Bruza, P., Gabora, L. (2012). A quantum information retrieval approach to memory. In press, Proceedings of the 2012 International Joint Conference on Neural Networks (IJCNN) to be held as a part of the World Congress on Computational Intelligence (WCCI).

% Koza, J. (1993). Genetic programming, MIT Press.

\item Klein, R. (1999). {\it The human career: Human biological and cultural origins.} Chicago: University of Chicago Press.

\item Leijnen, S.,  \& Gabora, L. (2009). How creative should creators be to optimize the evolution of ideas? A computational model. {\it Electronic Proceedings in Theoretical Computer Science, 9,} 108-119.

\item Leakey, R. (1984). {\it The origins of humankind}. New York: Science Masters Basic Books.

%% \item Leijnen, S.  \& Gabora, L. (2010). An agent-based simulation of the effectiveness of creative leadership. {\it Proceedings of the Annual Meeting of the Cognitive Science Society} (pp. 955-960). August 11-14, Portland, Oregon.

\item McBrearty, S., \& Brooks, A. S. (2000). The revolution that wasn�t: A new interpretation of the origin of modern human behavior. {\it Journal of Human Evolution, 39,} 453-563.

% \item Mesoudi, A., Whiten, A.,  \& Laland, K. (2006). Toward a unified science of cultural evolution. {\it Behavioral and Brain Sciences, 29,} 329-383.

% Miller, J. (2011). Cartesian genetic programming, Springer.

\item Mithen, S. (1998). {\it Creativity in human evolution and prehistory.} London, UK: Routledge.

% \item Osherson, D., & Smith, E. (1981). On the adequacy of prototype theory as a theory of concepts. Cognition, 9, 35 58. 

% Padian, K. (2008). Darwin's enduring legacy, Nature, 451, 632-634. 

% \item Premack, D. (1988). ``Does the chimpanzee have a theory of mind?" revisited. In R. W. Byrne \& A. Whiten (Eds.) {\it Machiavellian intelligence: Social expertise and the evolution of intellect in monkeys, apes and humans.} Oxford, UK: Oxford University Press.

% Ruff, C., Trinkaus, E., & Holliday, T. (1997. Body mass and encephalization in Pleistocene \emph{Homo}. Nature, 387, 173-176.

% Schilling, M. (2005). A "small-world" network model of cognitive insight. Creativity Research Journal, 17(2&3), 131-154.

\item Tomasello, M., Kruger, A.,  \& Ratner, H. (1993). Cultural learning. {\it Behavioral and Brain Sciences, 16,} 495-552.

% Ward, T. B., Smith, S. M.,  \& Vaid, J. (1997). Conceptual structures and processes in creative thought. In Ward, T. B., Smith, S. M. & Vaid. J. (Eds.) Creative Thought: An investigation of conceptual structures and processes (pp. 1-27). Washington, DC: American Psychological Association.

\item Veloz, T., Gabora, L., Eyjolfson, M. \& Aerts, D. (2011). A model of the shifting relationship between concepts and contexts in different modes of thought. {\it Proc 5th Intnl Symposium on Quantum Interaction.} Heidelberg: Springer. 
% Aberdeen, UK.

\item Wexler, B. (2006). {\it Brain and culture: Neurobiology, ideology and social change.} New York: Bradford Books.

% \item Whiten, A., Hinde, R., Laland, K.,  \& Stringer, C. (2011). Culture evolves. {\it Philosophical Transactions of the Royal Society B, 366,} 938-948.

\nocite{ChalnickBillman1988a}
\nocite{Feigenbaum1963a}
\nocite{Hill1983a}
\nocite{OhlssonLangley1985a}
\nocite{Lewis1978a}
\nocite{NewellSimon1972a}
\nocite{ShragerLangley1990a}

\bibliographystyle{apacite}

\setlength{\bibleftmargin}{.125in}
\setlength{\bibindent}{-\bibleftmargin}

\bibliography{CogSci_Template}

\end{description}
\end{document}